\documentclass[10pt,twocolumn,letterpaper]{article}

\usepackage{iccv}
\usepackage{times}
\usepackage{epsfig}
\usepackage{graphicx}
\usepackage{amsmath}
\usepackage{amssymb}
\usepackage{booktabs}
\usepackage{verbatim}
\usepackage{multirow}
\usepackage{makecell}
\usepackage{gensymb}
\usepackage{enumitem}

\usepackage[pagebackref=true,breaklinks=true,letterpaper=true,colorlinks,bookmarks=false]{hyperref}

\iccvfinalcopy 

\def\iccvPaperID{****} 

\ificcvfinal\fi

\usepackage[pagebackref=true,breaklinks=true,letterpaper=true,colorlinks,bookmarks=false]{hyperref}
\def\iccvPaperID{6656} 

\usepackage[capitalize]{cleveref}
\crefname{section}{Sec.}{Secs.}
\Crefname{section}{Section}{Sections}
\Crefname{table}{Table}{Tables}
\crefname{table}{Tab.}{Tabs.}
\usepackage{color}
\usepackage{colortbl}
\usepackage[table]{xcolor}
\definecolor{col1}{RGB}{243, 230, 195}
\definecolor{Gray}{gray}{0.9}
\definecolor{first_red}{RGB}{253, 132, 136}
\definecolor{second_orange}{RGB}{254, 193, 136}
\definecolor{third_yellow}{RGB}{254, 255, 136}
\newcolumntype{a}{>{\columncolor{Gray}}c}

\title{LivePose: Online 3D Reconstruction from Monocular Video with Dynamic Camera Poses}

\author{
Noah Stier$^{1,2}$ \hspace{3pt}
Baptiste Angles$^{1}$ \hspace{3pt}
Liang Yang$^{1}$ \hspace{3pt}
Yajie Yan$^{1}$ \hspace{3pt}
Alex Colburn$^{1}$ \hspace{3pt}
Ming Chuang$^{1}$
\\[1em]
$^1$Apple \hspace{6pt}
$^2$University of California, Santa Barbara \hspace{3pt}
}

\usepackage{overpic}

\newcommand{\rone}[1]{\textcolor[rgb]{1, 0, 0}{\textbf{R1}}}
\newcommand{\rtwo}[1]{\textcolor[rgb]{0, .75, 0}{\textbf{R2}}}
\newcommand{\rthree}[1]{\textcolor[rgb]{0, 0, 1}{\textbf{R3}}}

\newcommand{\integration}{\mathcal{F}^+}
\newcommand{\deintegration}{\mathcal{F}^-}
\newcommand{\image}{\mathbf{I}}
\newcommand{\pose}{\mathbf{T}}
\newcommand{\reconstruction}{\mathbf{R}}
\newcommand{\gt}{\hat{\reconstruction}}
\newcommand{\bundle}{\mathbf{B}}
\newcommand{\depth}{\mathbf{D}}

\newcommand{\firstcell}{\cellcolor{red!35}}
\newcommand{\secondcell}{\cellcolor{orange!35}}

\usepackage{cuted}
\usepackage{capt-of}

\begin{document}

\maketitle
\begin{strip}
    \centering
    \includegraphics[width=\textwidth]{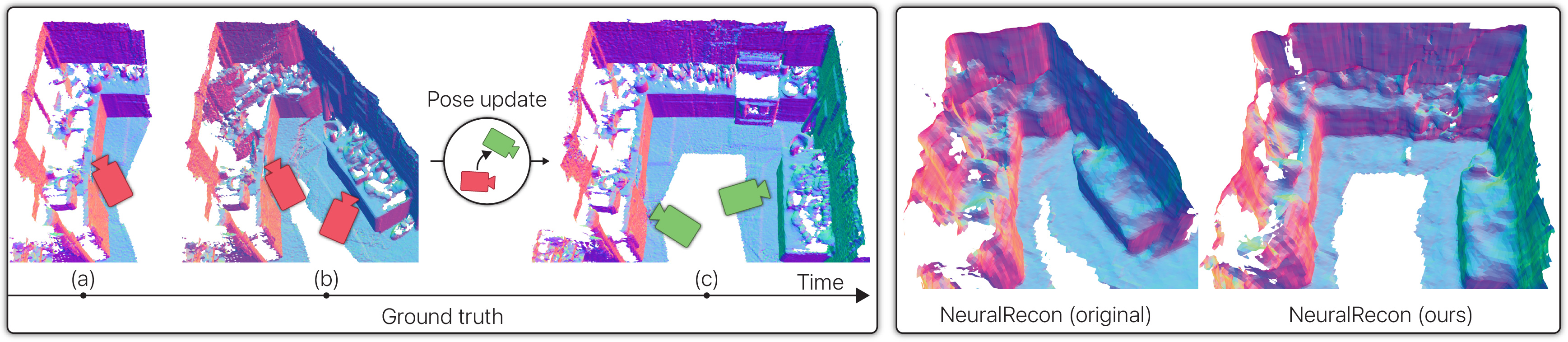}
    \captionof{figure}{
        In online 3D reconstruction, poses obtained from a SLAM system (a, b) may be updated (c, red$\rightarrow$green). Ignoring these updates leads to inconsistent geometry, while our pose update handling strategy produces accurate and globally consistent reconstructions.
        \label{fig:teaser}
    }
\vspace{-0.2em}
\end{strip}

\begin{abstract}
Dense 3D reconstruction from RGB images traditionally assumes static camera pose estimates. This assumption has endured, even as recent works have increasingly focused on real-time methods for mobile devices. However, the assumption of a fixed pose for each image does not hold for online execution: poses from real-time SLAM are dynamic and may be updated following events such as bundle adjustment and loop closure. This has been addressed in the RGB-D setting, by de-integrating past views and re-integrating them with updated poses, but it remains largely untreated in the RGB-only setting. We formalize this problem to define the new task of dense online reconstruction from dynamically-posed images. To support further research, we introduce a dataset called LivePose
\footnote{\href{https://github.com/apple/ml-live-pose}{https://github.com/apple/ml-live-pose}}
containing the dynamic poses from a SLAM system running on ScanNet \cite{scannet}. We select three recent reconstruction systems and apply a framework based on de-integration to adapt each one to the dynamic-pose setting. In addition, we propose a novel, non-linear de-integration module that learns to remove stale scene content. We show that responding to pose updates is critical for high-quality reconstruction, and that our de-integration framework is an effective solution.

\end{abstract}
\section{Introduction}
\label{sec:intro}

RGB-only reconstruction using a monocular camera has seen great progress using learned priors to address the difficulties associated with low-texture regions and the inherent ambiguity of image-based reconstruction. In particular, there has been strong interest in methods that are practical for real-time execution, a key component required for interactive applications on mobile devices. However, there is an additional requirement that has not been addressed in recent state-of-the-art reconstruction systems: a successful method must not only be real-time but also \textit{online}.

Online operation implies that an algorithm must produce accurate incremental reconstructions at the time of image acquisition, using only past and present observations at each time point. This problem setting violates a key assumption made by existing works: \textbf{the availability of an accurate, fully-optimized pose estimate for each view}. Instead, in a real-world scanning scenario, a Simultaneous Localization and Mapping (SLAM) system suffers pose drift, resulting in a stream of \textit{dynamic} pose estimates, where past poses are updated due to events such as pose graph optimization and loop closure. As detailed in Section \ref{sec:data}, such pose updates from SLAM are ubiquitous in online scanning. It is critical for the reconstruction to stay in agreement with the SLAM system by respecting these updates, as we have illustrated in Figure \ref{fig:teaser}.

Recent works on dense RGB-only reconstruction, however, have not addressed the dynamic nature of camera pose estimates in online applications \cite{transformerfusion,deepvideomvs,atlas,rich20213dvnet,vortx,neuralrecon}. Although these efforts have made great progress in reconstruction quality, they have preserved the traditional problem formulation of statically-posed input images, providing no explicit mechanism for handling dynamic poses.  In contrast, we acknowledge the presence of these updates, and we propose a solution by which existing RGB-only methods can incorporate pose update handling. We draw inspiration from BundleFusion \cite{bundlefusion}, an RGB-D method that integrates new views into the scene with a linear update rule, such that past views can be de-integrated and re-integrated when an updated pose becomes available. 

In this paper, we propose to use de-integration as a general framework for handling pose updates in online reconstruction from RGB images. We study three representative RGB-only reconstruction methods that have assumed static poses \cite{deepvideomvs,atlas,neuralrecon}. For each method, we apply the de-integration framework as detailed in Section \ref{sec:method} to address its limitations with respect to the online setting. In particular, we develop a novel, non-linear de-integration method based on deep learning to support online reconstruction for methods such as NeuralRecon \cite{neuralrecon} that use a learned, non-linear update rule. To validate this methodology and support future research, we release a novel and unique dataset called LivePose, featuring complete, dynamic pose sequences for ScanNet \cite{scannet}, generated using BundleFusion \cite{bundlefusion}. In our experiments (Section \ref{sec:experiments}) we demonstrate the effectiveness of the de-integration approach, showing qualitative and quantitative improvement among three state-of-the-art systems with respect to key reconstruction metrics.

\textbf{Contributions.} Our main contributions are as follows:

\setlist{nolistsep}
\begin{itemize}[noitemsep]
    \item We introduce and formalize a new vision task, dense online 3D reconstruction from dynamically-posed RGB images, that more closely reflects the real-world setting for interactive applications on mobile devices.
    \item We release LivePose: the first publicly-available dataset of dynamic SLAM pose estimates, containing the full SLAM pose stream for all 1,613 scans in the ScanNet dataset.
    \item We develop novel training and evaluation protocols to support reconstruction with dynamic poses.
    \item We propose a novel recurrent de-integration module that learns to handle pose updates by removing stale scene content, enabling dynamic-pose handling for methods with learned, recurrent view integration.
\end{itemize}
\section{Related Work}
\label{sec:related}

\begin{figure*}[t]
\begin{center}
    \begin{overpic}
    [width=\linewidth]
    {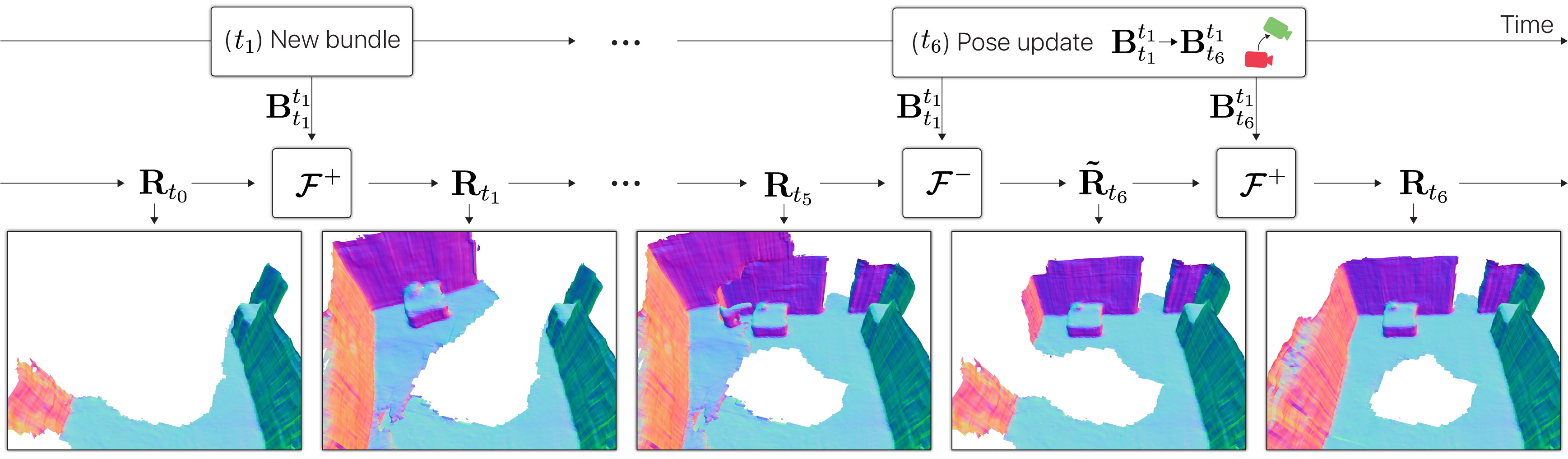}
    \end{overpic}
    \end{center}
    \vspace{-1em}
    \caption{
        \textbf{Pose update handling by de-integration.} 
        The integration module
        $\integration{}$incorporates a new bundle $\bundle_{t_1}^{t_1}$ into the scene $\reconstruction_{t_0}$ as it is received at time $t_1$. Later at $t_6$ when the SLAM system issues a pose update, the de-integration module $\deintegration{}$processes the same outdated bundle $\bundle_{t_1}^{t_1}$ to remove it from the scene. Then the updated bundle $\bundle_{t_6}^{t_1}$ with new pose estimates is re-integrated into the scene by $\integration{}$. \looseness=-1
    }
\label{fig:timeline}
\vspace{-0.8em}
\end{figure*}

Here we discuss the rationale for selecting representative reconstruction systems to evaluate using de-integration for live pose handling. Selected methods are in bold.

\subsection{Dense visual SLAM}

Although SLAM often uses additional hardware such as depth sensors \cite{schops2019bad, niceslam}, we instead focus on visual SLAM methods with no depth input, to support our use case of RGB-only reconstruction. Online mapping has been a long-standing goal in the visual SLAM literature, and a number of methods have evolved into mature, industry-standard software tools \cite{dsoslam,svoslam,orbslam}. A subset of visual SLAM methods also construct dense scene representations  \cite{bloesch2018codeslam,dsoslam,yang2020d3vo}, which could be used for downstream applications. These methods typically optimize jointly over the camera pose and dense depth for each input image. 
However, in our target application scenario, we assume that the poses are already estimated online, using an onboard SLAM system provided by the sensor platform.
This is a reasonable assumption for smartphones and other mobile devices, and it has two main benefits: 1) it eliminates the additional degrees of freedom and computational complexity that are incurred by optimizing jointly over camera pose, and 2) it renders our solutions agnostic to the particular choice of SLAM system.

\subsection{Depth-based reconstruction}

Image-based 3D reconstruction is traditionally posed as per-pixel depth estimation, followed by fusion into scene space to form a unified 3D model \cite{
deepvideomvs,furukawa2015multi,galliani2015massively,deepmvs,rich20213dvnet,simplerecon,colmap,mvsnet}. In particular, we select \textbf{DeepVideoMVS} \cite{deepvideomvs} as a representative method in this group due to its high accuracy and fast execution time. We also highlight
several notable limitations of reconstruction by depth estimation: 1) it provides no guarantee of agreement between different depth maps observing the same surfaces--this can result in incoherence in the fused model, which may require strong filtering or smoothing to resolve; 2) it cannot fill holes that arise in the reconstruction due to unobserved or occluded regions; 3) it does not offer a solution for end-to-end learning of 3D outputs for auxiliary tasks such as 3D semantic segmentation.

\subsection{Implicit volumetric reconstruction}

Recently, a number of methods have demonstrated a new paradigm in which a unified 3D model is estimated directly in scene space \cite{transformerfusion,choe2021volumefusion,feng2023cvrecon,atlas,vortx,neuralrecon}, alleviating the above limitations of depth prediction. These algorithms back-project deep feature maps extracted from the input images, forming a world-aligned feature volume over the scene. Then, they apply a 3D CNN to predict a truncated signed distance function (TSDF), and the result can be visualized by extracting an explicit mesh using marching cubes \cite{marchingcubes}. This approach guarantees multi-view consistency of the reconstruction, and it provides a natural framework for hole-filling and surface completion using learned 3D shape and scene composition priors. Moreover, these methods can be trained end-to-end for arbitrary volumetric prediction tasks, as demonstrated by Atlas \cite{atlas} for semantic segmentation.

\textbf{Atlas} \cite{atlas} integrates the features from each view into the scene volume using a linear running average. This approach lends itself to a straightforward solution based on linear de- and re-integration. One drawback of this method is that the linear integration function may result in reduced reconstruction quality relative to subsequent methods with learned, non-linear view integration. \textbf{NeuralRecon} \cite{neuralrecon} uses a recurrent neural network (RNN) to integrate new views, yielding improved reconstruction quality relative to integration based on unweighted averaging. However, the non-linear nature of this integration function also prevents the direct de-integration of arbitrary past observations, leaving NeuralRecon with no clear strategy for handling pose updates. VoRTX \cite{vortx} further demonstrates the quality improvements that can be attained with non-linear view integration, using transformers \cite{vaswani2017attention} to learn view selection and fusion. However, VoRTX provides no mechanism for producing incremental reconstructions as new views are added. The alternative is to reconstruct from scratch at each time point, which is prohibitively expensive for real-time applications.

TransformerFusion \cite{transformerfusion} also uses transformers to achieve increased reconstruction quality, and it introduces a view selection method that is compatible with incremental reconstruction. The proposed strategy is to maintain the top $N$ image features per voxel, fusing them via transformer at each time step.
However, storing $N$ features per voxel, instead of a single, accumulated feature, results in an $N$-fold increase in the memory footprint of the feature volume, as well as added computational cost due to repeatedly reducing across views. With the proposed $N=16$, these extra costs are significant, in particular when targeting mobile applications where power and memory are limited.
\section{Problem statement}
\label{sec:problem}

Given a sequence of images whose camera poses are estimated by an online SLAM system, we seek to compute an accurate online reconstruction of the scene.
As new observations are processed, camera pose estimates are continually updated by the SLAM system.
Formally, we denote the image captured at time $t_1$ as $\image^{t_1}$, and the estimate at time $t_2$ for the corresponding camera pose as $\pose_{t_2}^{t_1} \in \mathbb{SE}(3)$.

In a dynamic system, poses can change dramatically. Therefore, it is unreasonable to define the ground truth for time $t$ using the final pose estimates. 
Instead, the reconstruction system must respect the poses that are available at each point in time.
To define the desired output, we assume the existence of a ground-truth depth image $\depth^{t}$ paired with each color image $\image^{t}$.
The ground truth geometry at time $t$ is obtained by fusing all depths before and up to $t$:
\begin{equation}
    \gt_t = \operatorname{Fusion}(\{\depth^{t_i}\}, \{\pose_t^{t_i}\}), t_i \leq t,
\end{equation}
where $\operatorname{Fusion}$ represents the TSDF fusion operation\cite{curless1996volumetric}. 
\section{Method}
\label{sec:method}

\subsection{Generalized pose update handling}

Our solution to handle pose updates is a generalization of the strategy proposed by BundleFusion \cite{bundlefusion}.
Without loss of generality, we consider a reconstruction system that processes frame bundles incrementally. Depending on the specific system, such bundles can contain multiple views or just a single one.
We denote the bundle whose last image was captured at time $t_1$ and the corresponding pose estimates at time $t_2$ as $\bundle_{t_2}^{t_1}$:
\begin{equation}
    \bundle_{t_2}^{t_1} = \{\image^{t_i}, \pose_{t_2}^{t_i} \}_{t_i \in M^{1}},
\end{equation}
where $M^{1}$ is the set of image timestamps of the bundle created at time $t_1$.

When a new bundle is integrated, the reconstruction state produced by such a system at time $t$, denoted $\reconstruction_{t}$, is given by
\begin{equation}
    \reconstruction_{t} = \integration(\reconstruction_{t-1}, \bundle_{t}^{t}),
\end{equation}
where $\integration$ is the integration module which integrates a frame bundle to the previous reconstruction state $\reconstruction_{t-1}$.

When the pose estimates of a previously integrated bundle are updated by the SLAM system, we update the reconstruction state by first de-integrating the bundle's contribution and then integrating it back with the corrected pose.
Therefore, we define our strategy to update the poses of a bundle created at time $t_i$ and previously integrated at time $t_j$, with the new poses available at time $t$ as
\begin{equation}
        \reconstruction_{t} = \integration(\deintegration(\reconstruction_{t-1}, \bundle_{t_j}^{t_i}), \bundle_{t}^{t_i}),
\end{equation}
where $\deintegration$ refers to a de-integration operator which is responsible for un-doing the past integration of the bundle.
Our strategy is illustrated in Figure~\ref{fig:timeline}.

These two operators should be implemented such that $\deintegration$ is the inverse of $\integration$:
\begin{equation}
    \deintegration = (\integration)^{-1}
\end{equation}

The details of how $\reconstruction_t$, $\integration$ and $\deintegration$ are implemented vary depending on the method that is applied.
We discuss this topic for three classes of methods in the following sub-sections.

\subsection{Linear TSDF de-integration}
\label{sec:simplerecon}

The first reconstruction methodology we investigate is multi-view stereo (MVS) depth prediction. 
For this class of methods, we adopt a TSDF voxel grid for $\reconstruction_t$, and we use TSDF fusion to define our operators.
This gives the following definitions:
\begin{align}
    \integration(\reconstruction, \bundle) &= \operatorname{Fusion}(\reconstruction, \operatorname{MVS}(\bundle))\\
    \deintegration(\reconstruction, \bundle) &= \operatorname{Fusion}^{-1}(\reconstruction, \operatorname{MVS}(\bundle)),
\end{align}
where $\operatorname{Fusion}^{-1}$ is the TSDF fusion operator with a negative weight, and $\operatorname{MVS}$ is depth prediction network which leverages existing keyframes as auxiliary frames. These choices are preferred because the TSDF fusion operator is fast and $\deintegration$ is the exact inverse of $\integration$.

We choose DeepVideoMVS \cite{deepvideomvs} as a representative depth prediction method. 
\subsection{Linear de-integration for neural features}

We now turn to the direct, volumetric TSDF estimation approaches with linear integration.
In this paradigm, deep image features are extracted from each view by a network $\mathcal{E}$, and then densely back-projected into a voxel grid as $\reconstruction_t$, integrating features into each voxel via running average.
We observe that because this integration operation is linear, it supports de-integration similar to that of depth prediction methods, but operating on the neural features instead of directly on the TSDF values. Therefore, we define our operators using a running average:
\begin{align}
    \integration(\reconstruction, \bundle) &= \operatorname{Avg}^{+}(\reconstruction, \operatorname{BackProj}(\mathcal{E}(\bundle)))\\
    \deintegration(\reconstruction, \bundle) &= \operatorname{Avg}^{-}(\reconstruction, \operatorname{BackProj}(\mathcal{E}(\bundle)))
\end{align}

We use Atlas \cite{atlas} as a representative method in this category.

\subsection{Learned de-integration for neural features}
\label{sec:neuralrecon}

The third category of approach that we demonstrate is volumetric TSDF reconstruction with \textit{non}-linear multi-view fusion, as exemplified by NeuralRecon \cite{neuralrecon}.
In this case, the 2D deep features from $\mathcal{E}$ are integrated into the volume using a convolutional gated recurrent unit (GRU \cite{cho2014learning}) $\mathcal{G}^+$.

Recent trends show that there is a significant advantage to data-driven integration approaches \cite{transformerfusion, vortx, neuralrecon}, allowing the network to learn to attend to the most informative image inputs.
However, the recurrent, non-linear feature integration makes it intractable to de-integrate arbitrary past views because of the order-dependence.
We propose to address this by developing a novel, non-linear de-integration step, using deep learning to approximate the true de-integration function.
Our solution is to use a second convolutional GRU to serve as the de-integration network $\mathcal{G}^{-}$.
\begin{align}
    \integration(\reconstruction, \bundle) &= \mathcal{G}^{+}(\reconstruction, \operatorname{BackProj}(\mathcal{E}(\bundle)))\\
    \deintegration(\reconstruction, \bundle) &= \mathcal{G}^{-}(\reconstruction, \operatorname{BackProj}(\mathcal{E}(\bundle)))
\end{align}
The inputs to $\mathcal{G}^{-}$ are 1) the current scene feature volume and 2) the feature volume of the stale bundle to be de-integrated. We can interpret the desired function of this module in two steps. First, it must detect the effect that the target bundle has had on the current feature volume, and second, it must undo that effect.

We train our modified version of NeuralRecon on the LivePose dataset with the time-dependent ground truth $\hat{\mathbf{R}}_t$ defined in Section~\ref{sec:problem}.
This dynamic ground truth provides the necessary signal for learning the weights of the integration and de-integration networks, enabling the reconstruction system to remain in agreement with the SLAM system across updates.

Supervising the network with the data from LivePose has the added benefit that it exposes the network to a realistic distribution of online pose states, which may exhibit varying degrees of local and global consistency during online scanning. We show the impact of this effect in our ablation study (Table \ref{tab:ablation}, row (b) vs. (c)).

\begin{table*}[t]
\begin{center}
\begin{small}
\begin{tabular}{l c r r r r r r r r r}
    \toprule
     \multirow{2}{*}[-0.3em]{Base method} &  \multirow{2}{*}[-0.3em]{Integration} &  \multirow{2}{*}[-0.3em]{Strategy} &  \multirow{2}{*}[-0.3em]{Acc$\downarrow$} &  \multirow{2}{*}[-0.3em]{Comp$\downarrow$} &  \multirow{2}{*}[-0.3em]{Chamfer$\downarrow$} &  \multirow{2}{*}[-0.3em]{Prec$\uparrow$} &  \multirow{2}{*}[-0.3em]{Recall$\uparrow$} &  \multirow{2}{*}[-0.3em]{F-score$\uparrow$} & \multicolumn{2}{c}{Latency (ms)}\\
     \cmidrule(lr){10-11}
     &&&&&&&&&\multicolumn{1}{c}{New} & \multicolumn{1}{c}{Update} \\
    \midrule
    \multirow{3}{*}[0.0em]{Atlas \cite{atlas}} & \multirow{3}{*}{\makecell{Linear \\ (features)}} & No updates & 0.084 & \secondcell{0.183} & \secondcell{0.133} & 0.584 & 0.489 & 0.531 & 2,481 & 0 \\
     & & Re-integration & \secondcell{0.082} & 0.188 & 0.135 & \secondcell{0.602} & \secondcell{0.500} & \secondcell{0.545} & 2,492 & 2,363 \\
     & & \textbf{Ours} & \firstcell{0.078} & \firstcell{0.177} & \firstcell{0.127} & \firstcell{0.619} & \firstcell{0.520} & \firstcell{0.564} & 2,475 & 2,673 \\
    \midrule[0.1pt]
    \multirow{3}{*}[0.0em]{NeuralRecon \cite{neuralrecon}} & \multirow{3}{*}{\makecell{Non-linear \\ (features)}} & No updates & 0.056 & 0.215 & 0.136 & 0.667 & 0.469 & 0.546 & 223 & 0 \\
     & & Re-integration & \secondcell{0.053} & \firstcell{0.204} &  \firstcell{0.129} & \secondcell{0.688} & \secondcell{0.492} & \secondcell{0.568} & 223 & 233 \\
     & & \textbf{Ours} & \firstcell{0.048} & \secondcell{0.211} & \firstcell{0.129} & \firstcell{0.708} & \firstcell{0.495} & \firstcell{0.577} & 211 & 426 \\
    \midrule[0.1pt]
    \multirow{3}{*}[0.0em]{DeepVideoMVS \cite{deepvideomvs}} & \multirow{3}{*}{\makecell{Linear \\ (TSDF)}} & No updates & 0.082 & 0.096 & 0.089 & 0.536 & 0.508 & 0.521 & 1,081 & 0 \\
     & & Re-integration & \secondcell{0.080} & \secondcell{0.092} & \secondcell{0.086} & \secondcell{0.550} & \secondcell{0.524} & \secondcell{0.535} & 1,081 & 143 \\
     & & \textbf{Ours} & \firstcell{0.071} & \firstcell{0.086} &\firstcell{0.079} & \firstcell{0.569} & \firstcell{0.543} & \firstcell{0.554} & 1,081 & 287 \\
    \bottomrule
\end{tabular}
\end{small}
\end{center}
\caption{\textbf{Online 3D reconstruction metrics for ScanNet}, using the live pose data from LivePose. Comparison with state-of-the-art methods using the reconstruction metrics defined in Murez et al. \cite{atlas}, where the Chamfer distance is the mean of accuracy and completeness. Note that we report the latency per bundle (nine keyframes).
}
\label{tab:comparison}
\vspace{-0.2em}
\end{table*}

\subsection{Pose update filtering}
\label{sec:bundles}
We observe that the reconstruction systems investigated in this paper require a non-negligible amount of time to perform an update (see Table \ref{tab:comparison}). 
As a result, it is not tractable to keep the reconstruction in perfect synchrony with a SLAM system that provides high-frequency pose updates for a large number of past views.
We therefore propose that an intermediate layer should filter the pose stream from SLAM, emitting an update to the reconstruction system each time that the current optimized poses have drifted sufficiently far from their previously-integrated state.
This strategy allows us to ensure a tractable pose update frequency regardless of the choice of SLAM implementation.

We propose to fill the role of the intermediate layer using the fragment generation system from NeuralRecon \cite{neuralrecon}, which we extend in order to incorporate pose update dynamics. 
The first step in this system is to sub-sample a set of keyframes from the SLAM pose stream, adding a keyframe each time that the current camera pose differs from the previous keyframe by at least 10cm of translation or 15\degree of rotation. 
This supports real-time processing by reducing the total number of frames to be processed. 
Secondly, each time that $K$ new keyframes are available, we collect them into a frame bundle, that is issued to the reconstruction system, thus increasing efficiency by batch processing. Following NeuralRecon, we set $K=9$.

When the sum of the update distances over all views in bundle crosses a threshold $d$, we produce two additional bundles.
The first is a de-integration bundle containing the now-stale poses that should be removed from the reconstruction. 
The second is a re-integration bundle containing the new optimized poses for the views in question. We set $d = 0.45$m, corresponding to an average update distance of $5$cm over each of the nine keyframes.

\subsection{Memory cost}

For DeepVideoMVS, we require storing a depth image for each keyframe, which is tractable using our keyframe selection parameters. For Atlas and NeuralRecon, we require storing a feature map for each keyframe, although this space cost could be traded for latency by instead storing RGB images, and recomputing the feature maps when needed. The de-integration GRU in NeuralRecon incurs the memory cost of storing and running one extra neural network. However, it crucially does not require storing a second feature volume as it shares the scene state with the integration GRU.
\section{LivePose dataset}
\label{sec:data}

\begin{figure}[b]
\vspace{-2em}
\begin{center}
    \begin{overpic}
    [width=0.8\linewidth]
    {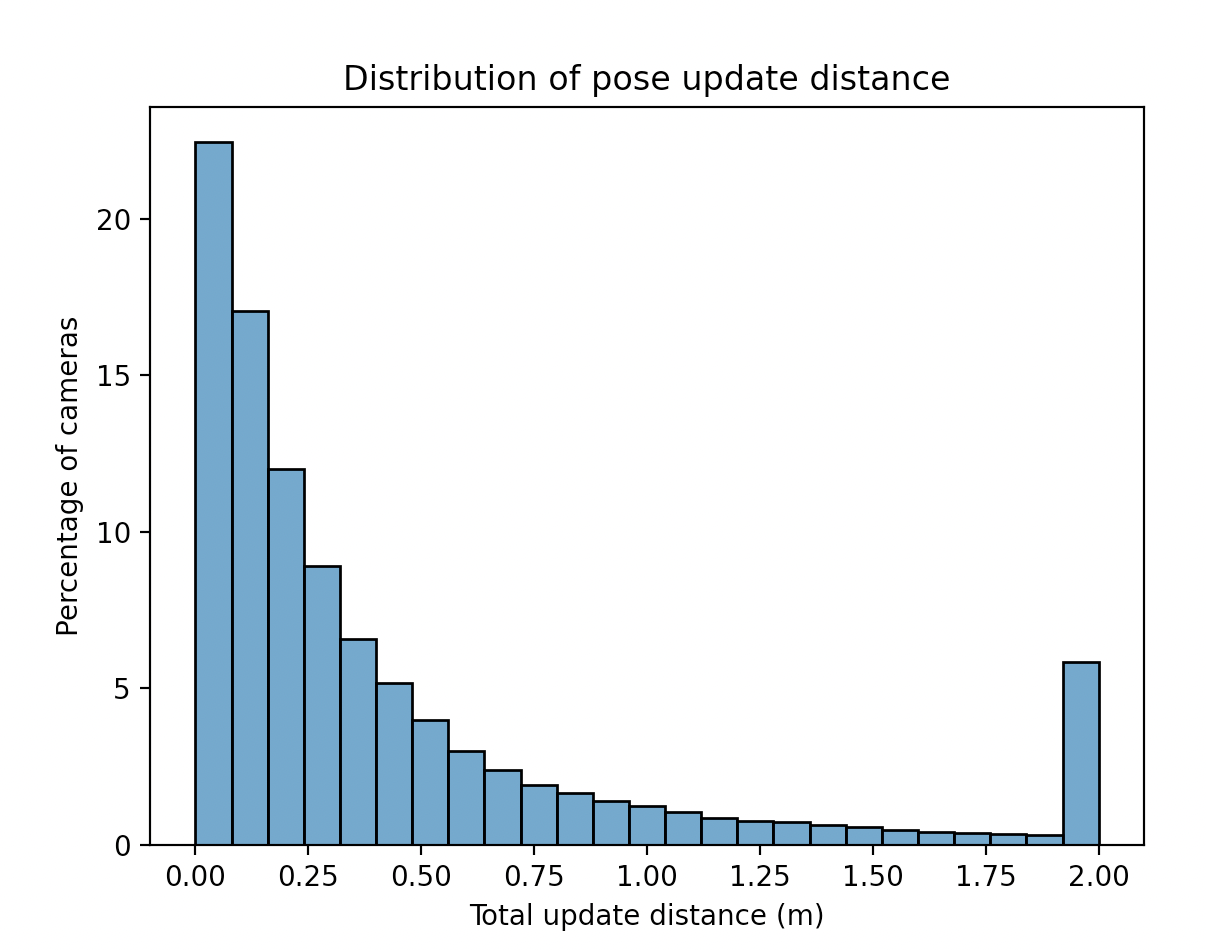}
    \end{overpic}
    \end{center}
    \vspace{-1em}
    \caption{Histogram of the total update distance traveled by each camera as its pose estimate changes throughout the scan. Values are clipped to a maximum of $2$m. Large pose updates are common with over $10$\% of poses being displaced by at least $1$m.}
\label{fig:LivePoseHist}
\end{figure}

In order to develop and evaluate our proposed de-integration strategies, we require access to the dynamic SLAM pose stream for a large number of RGB-D scans. However, to the best of our knowledge, there is no publicly available dataset that meets this need. The ScanNet dataset \cite{scannet} contains 1,613 RGB-D scans of indoor scenes, and while it has played a significant role in recent 3D reconstruction research, its public release includes only the final, fully-optimized pose estimates for each scan.

Therefore, in order to utilize ScanNet in our dynamic-pose setting, we estimate the full, dynamic pose sequence for each scan. For consistency with the original ScanNet release, we use BundleFusion \cite{bundlefusion} as the SLAM system. In the resulting live pose streams, updates are highly prevalent: 
\begin{itemize}
    \item \textbf{98\% of pose estimates receive at least one update}
    \item \textbf{25\% of pose estimates are displaced by at least 0.5m}
    \item \textbf{5\% of pose estimates are displaced by at least 2m}
\end{itemize}
See Figure \ref{fig:LivePoseHist} for more detailed statistics.

\begin{figure*}
\begin{center}
    \begin{overpic}
    [width=0.95\linewidth]
    {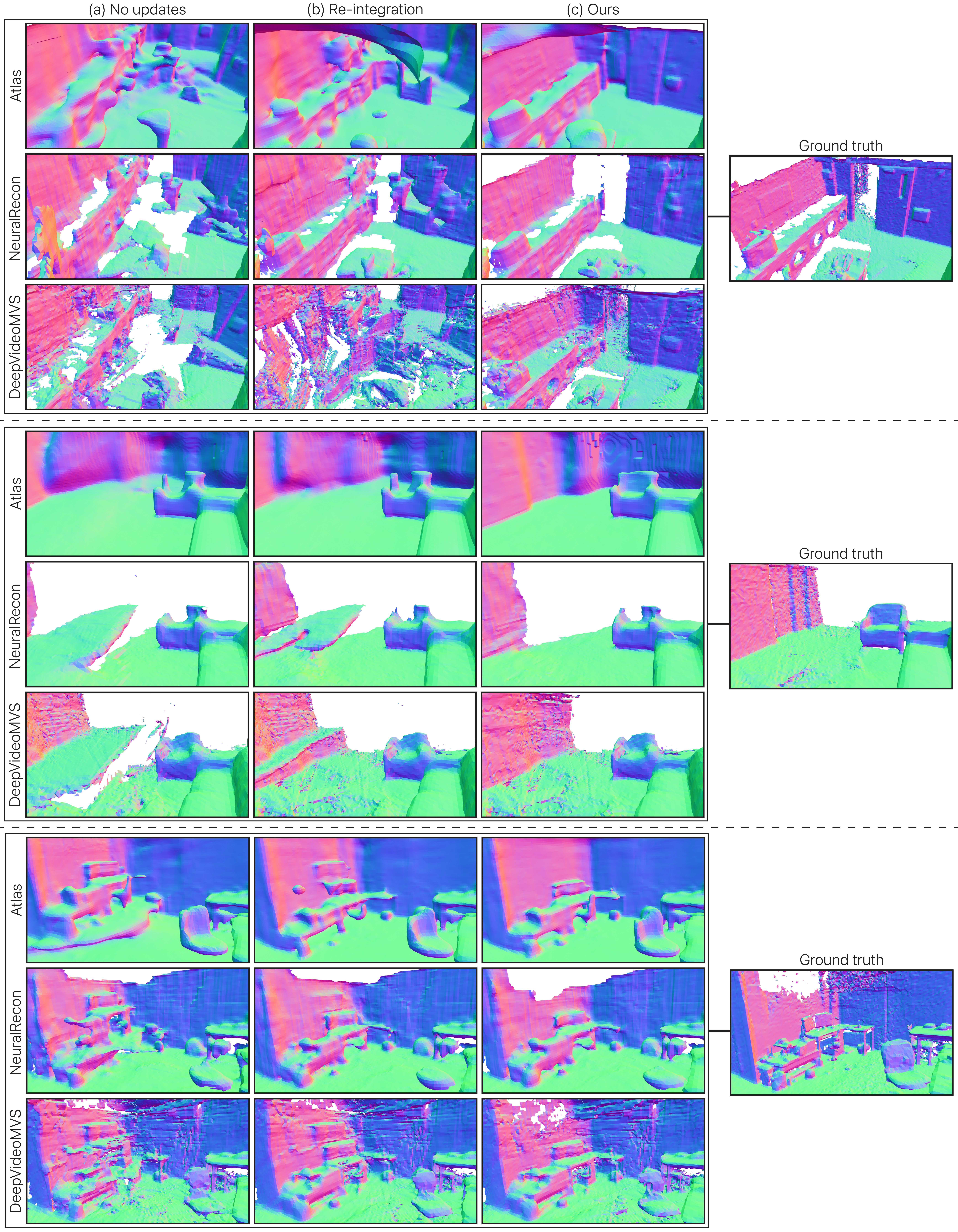}
    \end{overpic}
    \end{center}
    \vspace{-1em}
    \caption{
        \textbf{Qualitative results with LivePose.} 
        Camera poses from real-time SLAM can exhibit significant drift, leading to incoherent reconstruction states. (a) When pose updates are issued, failing to respond to them leads to significant discrepancies with respect to the ground truth reconstruction. (b) Re-integrating with updated poses can recover missing geometry, but (c) de-integration is necessary in order to remove stale scene content and achieve a coherent reconstruction.
        }
\label{fig:qualitative}
\end{figure*}

We register our estimated trajectories to the official ground-truth poses following Horn \cite{horn1987closed}, to facilitate comparison and to ensure that the gravitational axis is oriented consistently across scenes. To validate the accuracy of our pose data, we compare our BundleFusion pose estimates at the final time step of each scene to the official ground-truth poses, after registration. On average, we observe a $5$cm mean absolute error (MAE) in camera position over all scenes. Upon qualitative investigation, we find that scenes with less than $15$cm MAE typically do not have visually salient tracking errors, and our pose estimates fall below this threshold in $97\%$ of scenes. In our dynamic pose data, we have valid pose estimates for $97.7\%$ of total frames, compared to $97.9\%$ in the official ground truth. 

We release the dynamic pose sequences in the form of the publicly available LivePose dataset. This data is a key component of our training and evaluation, and we hope it will be a useful tool and point of reference for future work.

\section{Experiments}
\label{sec:experiments}

We validate our solutions on ScanNet with dynamic pose estimates from LivePose, using the 100 scenes of the official test set to compute reconstruction metrics. In the tables, {\colorbox{red!35}{Best}} and \colorbox{orange!35}{Second}-best are highlighted. 

\subsection{Baselines}

For Atlas and DeepVideoMVS, we use the ScanNet-trained weights provided by the original authors in all experiments. For NeuralRecon, we re-train to incorporate our learned de-integration module, and for consistency we therefore use our trained weights for the NeuralRecon baselines. For DeepVideoMVS, we relax the pose update strategy by computing the depth for each view only once, and de-/re-integrating that same depth when the poses are updated. This reduces the update time significantly, and we observe no loss in accuracy due to high internal pose consistency within bundles across updates.

\vspace*{-1.0em}
\paragraph{No updates}

To demonstrate the impact of failing to respond to pose updates, we evaluate each method without any modifications for handling dynamic pose. In this setting, the input sequence consists of each new bundle at the time that it first becomes available, with no subsequent updates. However, the ground-truth sequence does receive updates, and this allows us to quantify the discrepancy created between the SLAM system and the reconstruction system.

\vspace*{-1.0em}
\paragraph{Re-integration only}

As a naive baseline for handling pose updates, we perform re-integration at each update without first performing de-integration.
Stale scene content thus lingers in the reconstruction, and the results allow us to directly observe the contribution of de-integration.

\subsection{Evaluating with dynamic poses}

Previous works have not used dynamic poses for evaluation. In addition, they have not evaluated the quality of the incremental reconstructions at intermediate time steps.
One challenge in performing this evaluation is that the official ScanNet data only contains ground-truth reconstructions for the final time step of each sequence. A second challenge is that in order to compare across methods, each one must produce reconstructions at the same set of fixed time points.

We address these challenges by first generating and publishing the dynamic pose sequences of LivePose. We then define a set of common, incremental reconstruction checkpoints: we set a checkpoint for every time step at which a bundle is integrated. This naturally produces a variable-rate series of checkpoints, more frequently sampling the reconstruction during periods of rapid updates. We generate a ground-truth mesh for each checkpoint by online TSDF fusion and marching cubes \cite{marchingcubes}, at a resolution of $4$cm.

\subsection{Pose update handling by de-integration}

\subsubsection{3D reconstruction evaluation}
Figure \ref{fig:qualitative} shows qualitative results, using time points chosen before and after a series of pose updates. These results illustrate the significant incoherence that can result from pose drift, as well as the ability of our de-integration solutions to remain in agreement with the SLAM system as it recovers. We also observe that re-integration alone, without de-integration, is not enough to remove the stale scene content.

As shown in Table \ref{tab:comparison}, de-integration leads to the best outcome on all metrics. This supports our claim that direct de-integration is a practical, effective framework for handling pose updates, and we show how it can be generalized across multiple scene representations and integration strategies.

\subsubsection{Online reconstruction efficiency}
Timing results are computed as an average over the ScanNet test set on an NVIDIA V100 GPU (Table \ref{tab:comparison}).
Atlas averages 2,673ms to process an update event for a bundle, comparable to its time for new bundle integration. In the case of NeuralRecon, handling an update requires running the 3D CNN twice, thus doubling the execution time for updates relative to new bundles. However, each update still requires less than 500ms, which is permissible in many interactive applications. For DeepVideoMVS, the update time is much faster than the new bundle integration time, primarily because we opt not to re-compute depth for each update; the timing breakdown for a processing a new bundle (1,081ms total) is on average 104ms for depth estimation and 16ms for TSDF fusion, for each of the nine frames.

\subsection{Ablation studies}
\label{sec:ablations}

\begin{table}
\begin{center}
\resizebox{\linewidth}{!}{
\small
\begin{tabular}{l @{\hspace{3pt}} l c r r}
    \toprule
    & \multirow{2}{*}[-0.3em]{Method} & \multirow{2}{*}[-0.3em]{F-score$\uparrow$} & \multicolumn{2}{c}{Latency (ms)} \\
    \cmidrule(lr){4-5}
    & & & \multicolumn{1}{c}{New} & \multicolumn{1}{c}{Update}\\
    \midrule
    & \textbf{NeuralRecon \cite{neuralrecon}} & & & \\
    \midrule[0.1pt]
    (a) & Learned de-integration (ours) & \secondcell{0.577} & 211 & 426 \\
    (b) & Reconstruct from scratch & \firstcell{0.581} & 233 & 5,487 \\
    (c) & Reconstruct from scratch (authors' weights) & 0.544 & 204 & 4,724 \\
    \midrule
    & \textbf{DeepVideoMVS \cite{deepvideomvs}} with our de-integration & & & \\
    \midrule[0.1pt]
    (d) & Fusion network & \firstcell{0.554} & 1,031 & 287 \\
    (e) & Pair network & 0.515 & 919 & 284 \\
    (f) & Fusion network (re-compute depth) & \secondcell{0.545} & 1,031 & 1,233 \\
    \bottomrule
\end{tabular}
}
\end{center}
\caption{\textbf{Ablation study.} 
For NeuralRecon, we compare our learned de-integration (a) with an upper bound obtained by reconstruction from scratch for each update (b). In (c) we repeat (b) using the model weights provided by the original authors, showing the positive impact of our training protocol with live ground truth TSDF. For DeepVideoMVS, we show the effects of using the pair network (e), and re-computing depth before re-integration (f). See Section \ref{sec:ablations} for details and discussion.
}
\label{tab:ablation}
\vspace{-1em}
\end{table}

\begin{figure*}[t]
\begin{center}
    \begin{overpic}
    [width=\linewidth]
    {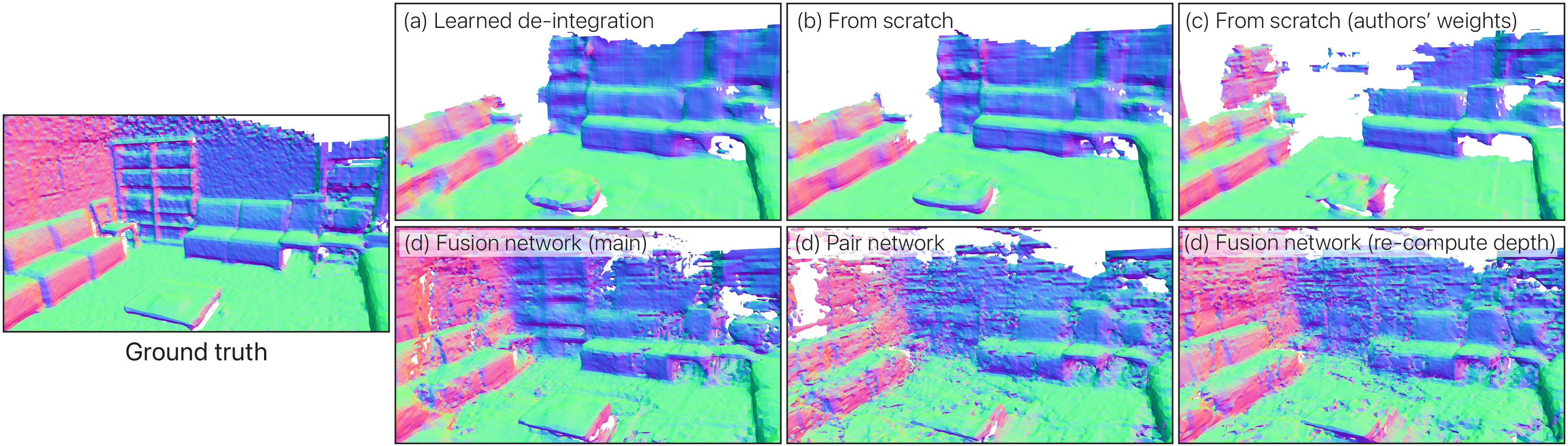}
    \end{overpic}
    \end{center}
    \vspace{-1em}
    \caption{
    \textbf{Visual results for ablation study (Table~\ref{tab:ablation}).}
    For NeuralRecon, our solution (a) approaches the quality of our upper bound based on reconstructing from scratch (b). The improvement from (c) to (b) shows the benefit of live-pose training. For DeepVideoMVS, (d, e) show results from the fusion vs. pair network, and (f) shows the effect of recomputing depth upon re-integration. Details in Section~\ref{sec:ablations}.
    }
\label{fig:ablations}
\end{figure*}

We further characterize our results using a series of ablation studies, with metrics reported in Table \ref{tab:ablation} and qualitative results shown in Figure \ref{fig:ablations}.

\textbf{Are we close to optimal de-integration?} In Table \ref{tab:ablation} (b-c) we simulate perfect non-linear de-integration performance by starting the reconstruction from scratch each time an update is received. This strategy requires over $6$s to process an update on average. While it is not a practical solution, it provides a useful upper bound on the accuracy of the de-integration module. Our ablation results are illustrated in Table \ref{tab:ablation}, where (a) represents our novel solution for learned, non-linear de-integration and (b) is the perfect de-integration. In (c) we repeat the perfect de-integration experiment using the NeuralRecon weights provided by the original authors, instead of our re-trained weights. The gain in F-score from (c) to (a) indicates the importance of training with dynamic poses, which helps the network to perform accurate reconstruction when the pose estimates have not yet been fully optimized. These results show that the learned de-integration achieves the fastest processing rate while maintaining comparable reconstruction quality.

\textbf{Should we re-compute depth?}
For DeepVideoMVS, updating the pose involves more than just integrating and de-integrating the per-view geometry. It also requires a re-evaluation of the depth computation due to potential changes in inter-frame relative poses. We perform experiments with vs without re-computing depth upon updates, and results are presented in the lower part of Table \ref{tab:ablation}. Row (d) shows our main result for DeepVideoMVS with de-integration. In (e) we use the pair network instead of the fusion network, showing a slight reduction in F-score, consistent with our findings. In (f) we additionally re-compute the depth for each reference view before each re-integration. In theory, this could be necessary in the case of significant relative pose updates within the bundle, but in practice, we find that the results are similar, indicating high internal pose consistency across updates.
\section{Conclusions and future work}
We have presented a new problem formulation for online monocular 3D reconstruction, in which a dense reconstruction system must account for pose update events such as loop closures issued by the SLAM system, and we have demonstrated the importance of handling these updates to create accurate reconstructions. To address this problem, we have developed and validated solutions based on de-integration for three representative state-of-the-art methods, and we have introduced a novel de-integration module that enables pose update handling for methods using non-linear view integration. We have prepared a unique and novel dataset of online SLAM pose estimates, which we hope will facilitate future research in this area.

{
    \small
    \bibliographystyle{ieee_fullname}
    \bibliography{references}

\begin{thebibliography}{10}\itemsep=-1pt

\bibitem{bloesch2018codeslam}
Michael Bloesch, Jan Czarnowski, Ronald Clark, Stefan Leutenegger, and Andrew~J
  Davison.
\newblock Codeslam—learning a compact, optimisable representation for dense
  visual slam.
\newblock In {\em Proceedings of the IEEE conference on computer vision and
  pattern recognition}, pages 2560--2568, 2018.

\bibitem{transformerfusion}
Alja{\v{z}} Bo{\v{z}}i{\v{c}}, Pablo Palafox, Justus Thies, Angela Dai, and
  Matthias Nie{\ss}ner.
\newblock Transformerfusion: Monocular rgb scene reconstruction using
  transformers.
\newblock {\em Proc. Neural Information Processing Systems (NeurIPS)}, 2021.

\bibitem{cho2014learning}
Kyunghyun Cho, Bart Van~Merri{\"e}nboer, Caglar Gulcehre, Dzmitry Bahdanau,
  Fethi Bougares, Holger Schwenk, and Yoshua Bengio.
\newblock Learning phrase representations using rnn encoder-decoder for
  statistical machine translation.
\newblock {\em arXiv preprint arXiv:1406.1078}, 2014.

\bibitem{choe2021volumefusion}
Jaesung Choe, Sunghoon Im, Francois Rameau, Minjun Kang, and In~So Kweon.
\newblock Volumefusion: Deep depth fusion for 3d scene reconstruction.
\newblock In {\em Proceedings of the IEEE/CVF International Conference on
  Computer Vision}, pages 16086--16095, 2021.

\bibitem{curless1996volumetric}
Brian Curless and Marc Levoy.
\newblock A volumetric method for building complex models from range images.
\newblock In {\em Proceedings of the 23rd annual conference on Computer
  graphics and interactive techniques}, pages 303--312, 1996.

\bibitem{scannet}
Angela Dai, Angel~X. Chang, Manolis Savva, Maciej Halber, Thomas Funkhouser,
  and Matthias Nie{\ss}ner.
\newblock Scannet: Richly-annotated 3d reconstructions of indoor scenes.
\newblock In {\em Proc. Computer Vision and Pattern Recognition (CVPR), IEEE},
  2017.

\bibitem{bundlefusion}
Angela Dai, Matthias Nie{\ss}ner, Michael Zoll{\"o}fer, Shahram Izadi, and
  Christian Theobalt.
\newblock Bundlefusion: Real-time globally consistent 3d reconstruction using
  on-the-fly surface re-integration.
\newblock {\em ACM Transactions on Graphics 2017 (TOG)}, 2017.

\bibitem{deepvideomvs}
Arda Duzceker, Silvano Galliani, Christoph Vogel, Pablo Speciale, Mihai
  Dusmanu, and Marc Pollefeys.
\newblock Deepvideomvs: Multi-view stereo on video with recurrent
  spatio-temporal fusion.
\newblock In {\em Proceedings of the IEEE/CVF Conference on Computer Vision and
  Pattern Recognition}, pages 15324--15333, 2021.

\bibitem{dsoslam}
Jakob Engel, Vladlen Koltun, and Daniel Cremers.
\newblock Direct sparse odometry.
\newblock {\em IEEE transactions on pattern analysis and machine intelligence},
  40(3):611--625, 2017.

\bibitem{feng2023cvrecon}
Ziyue Feng, Leon Yang, Pengsheng Guo, and Bing Li.
\newblock {CVRecon}: Rethinking 3d geometric feature learning for neural
  reconstruction.
\newblock In {\em Proceedings of the IEEE/CVF International Conference on
  Computer Vision}, 2023.

\bibitem{svoslam}
Christian Forster, Zichao Zhang, Michael Gassner, Manuel Werlberger, and Davide
  Scaramuzza.
\newblock Svo: Semidirect visual odometry for monocular and multicamera
  systems.
\newblock {\em IEEE Transactions on Robotics}, 33(2):249--265, 2016.

\bibitem{furukawa2015multi}
Yasutaka Furukawa, Carlos Hern{\'a}ndez, et~al.
\newblock Multi-view stereo: A tutorial.
\newblock {\em Foundations and Trends{\textregistered} in Computer Graphics and
  Vision}, 9(1-2):1--148, 2015.

\bibitem{galliani2015massively}
Silvano Galliani, Katrin Lasinger, and Konrad Schindler.
\newblock Massively parallel multiview stereopsis by surface normal diffusion.
\newblock In {\em Proceedings of the IEEE International Conference on Computer
  Vision}, pages 873--881, 2015.

\bibitem{horn1987closed}
Berthold~KP Horn.
\newblock Closed-form solution of absolute orientation using unit quaternions.
\newblock {\em Josa a}, 4(4):629--642, 1987.

\bibitem{deepmvs}
Po-Han Huang, Kevin Matzen, Johannes Kopf, Narendra Ahuja, and Jia-Bin Huang.
\newblock Deepmvs: Learning multi-view stereopsis.
\newblock In {\em Proceedings of the IEEE Conference on Computer Vision and
  Pattern Recognition (CVPR)}, June 2018.

\bibitem{marchingcubes}
William~E Lorensen and Harvey~E Cline.
\newblock Marching cubes: A high resolution 3d surface construction algorithm.
\newblock {\em ACM siggraph computer graphics}, 21(4):163--169, 1987.

\bibitem{orbslam}
Raul Mur-Artal, Jose Maria~Martinez Montiel, and Juan~D Tardos.
\newblock Orb-slam: a versatile and accurate monocular slam system.
\newblock {\em IEEE transactions on robotics}, 31(5):1147--1163, 2015.

\bibitem{atlas}
Zak Murez, Tarrence~van As, James Bartolozzi, Ayan Sinha, Vijay Badrinarayanan,
  and Andrew Rabinovich.
\newblock Atlas: End-to-end 3d scene reconstruction from posed images.
\newblock In {\em European conference on computer vision}, pages 414--431.
  Springer, 2020.

\bibitem{rich20213dvnet}
Alexander Rich, Noah Stier, Pradeep Sen, and Tobias H{\"o}llerer.
\newblock 3dvnet: Multi-view depth prediction and volumetric refinement.
\newblock In {\em 2021 International Conference on 3D Vision (3DV)}, pages
  700--709. IEEE, 2021.

\bibitem{simplerecon}
Mohamed Sayed, John Gibson, Jamie Watson, Victor Prisacariu, Michael Firman,
  and Cl{\'e}ment Godard.
\newblock Simplerecon: 3d reconstruction without 3d convolutions.
\newblock In {\em Proceedings of the European Conference on Computer Vision
  (ECCV)}, 2022.

\bibitem{colmap}
Johannes~L Sch{\"o}nberger, Enliang Zheng, Jan-Michael Frahm, and Marc
  Pollefeys.
\newblock Pixelwise view selection for unstructured multi-view stereo.
\newblock In {\em European conference on computer vision}, pages 501--518.
  Springer, 2016.

\bibitem{schops2019bad}
Thomas Schops, Torsten Sattler, and Marc Pollefeys.
\newblock Bad slam: Bundle adjusted direct rgb-d slam.
\newblock In {\em Proceedings of the IEEE/CVF Conference on Computer Vision and
  Pattern Recognition}, pages 134--144, 2019.

\bibitem{vortx}
Noah Stier, Alexander Rich, Pradeep Sen, and Tobias H{\"o}llerer.
\newblock Vortx: Volumetric 3d reconstruction with transformers for voxelwise
  view selection and fusion.
\newblock In {\em 2021 International Conference on 3D Vision (3DV)}, pages
  320--330. IEEE, 2021.

\bibitem{neuralrecon}
Jiaming Sun, Yiming Xie, Linghao Chen, Xiaowei Zhou, and Hujun Bao.
\newblock Neuralrecon: Real-time coherent 3d reconstruction from monocular
  video.
\newblock In {\em Proceedings of the IEEE/CVF Conference on Computer Vision and
  Pattern Recognition}, pages 15598--15607, 2021.
\newblock zju3dv.github.io/neuralrecon.

\bibitem{vaswani2017attention}
Ashish Vaswani, Noam Shazeer, Niki Parmar, Jakob Uszkoreit, Llion Jones,
  Aidan~N Gomez, {\L}ukasz Kaiser, and Illia Polosukhin.
\newblock Attention is all you need.
\newblock {\em Advances in neural information processing systems}, 30, 2017.

\bibitem{yang2020d3vo}
Nan Yang, Lukas~von Stumberg, Rui Wang, and Daniel Cremers.
\newblock D3vo: Deep depth, deep pose and deep uncertainty for monocular visual
  odometry.
\newblock In {\em Proceedings of the IEEE/CVF Conference on Computer Vision and
  Pattern Recognition}, pages 1281--1292, 2020.

\bibitem{mvsnet}
Yao Yao, Zixin Luo, Shiwei Li, Tian Fang, and Long Quan.
\newblock Mvsnet: Depth inference for unstructured multi-view stereo.
\newblock In {\em Proceedings of the European Conference on Computer Vision
  (ECCV)}, September 2018.

\bibitem{niceslam}
Zihan Zhu, Songyou Peng, Viktor Larsson, Weiwei Xu, Hujun Bao, Zhaopeng Cui,
  Martin~R Oswald, and Marc Pollefeys.
\newblock Nice-slam: Neural implicit scalable encoding for slam.
\newblock In {\em Proceedings of the IEEE/CVF Conference on Computer Vision and
  Pattern Recognition}, pages 12786--12796, 2022.

\end{thebibliography}
}

\end{document}